\documentclass[12pt]{article}
\usepackage{geometry}
\usepackage{graphicx}
\usepackage{amsmath, amssymb}
\usepackage{hyperref}

\usepackage{titling} 
\patchcmd{\maketitle}{plain}{empty}{}{} 

\title{\textbf{Organizing a Society of Language Models: Structures and Mechanisms for Enhanced Collective Intelligence}}
\author{%
    Silvan Ferreira\textsuperscript{1}\thanks{Corresponding author: \texttt{silvan.junior.051@ufrn.edu.br}}, Ivanovitch Silva\textsuperscript{2}, and Allan Martins\textsuperscript{3}
}

\date{\footnotesize{Graduate Program in Electrical and Computer Engineering (PPgEEC)\\ Federal University of Rio Grande do Norte (UFRN)}}

\begin{document}

\maketitle

\begin{abstract}
Recent developments in Large Language Models (LLMs) have significantly expanded their applications across various domains. However, the effectiveness of LLMs is often constrained when operating individually in complex environments. This paper introduces a transformative approach by organizing LLMs into community-based structures, aimed at enhancing their collective intelligence and problem-solving capabilities. We investigate different organizational models—hierarchical, flat, dynamic, and federated—each presenting unique benefits and challenges for collaborative AI systems. Within these structured communities, LLMs are designed to specialize in distinct cognitive tasks, employ advanced interaction mechanisms such as direct communication, voting systems, and market-based approaches, and dynamically adjust their governance structures to meet changing demands. The implementation of such communities holds substantial promise for improve problem-solving capabilities in AI, prompting an in-depth examination of their ethical considerations, management strategies, and scalability potential. This position paper seeks to lay the groundwork for future research, advocating a paradigm shift from isolated to synergistic operational frameworks in AI research and application.
\end{abstract}

\section{Introduction}
\label{sec:introduction}
As artificial intelligence continues to rapidly advance, Large Language Models (LLMs) like GPT and BERT have profoundly transformed computational capabilities, extending their applications from basic text processing to intricate decision-making across various sectors such as healthcare, finance, and public policy \cite{brown2020language, devlin2019bert}. These models, which are fundamentally based on the Transformer architecture, have considerably broadened the scope of achievable tasks through natural language processing \cite{vaswani2017attention}. However, despite their advanced capabilities, LLMs encounter significant challenges when tasked with complex, nuanced scenarios that demand a deep understanding of context and relational data, often resulting in less than optimal performance \cite{wei2022chain, berglund2024, sun2023think, reese2023limitations, ishay2023leveraging}. The integration of external knowledge bases and enhanced memory states have been proposed as potential solutions to mitigate these limitations \cite{sun2023think, statler2023state}.

Recent innovations in the field, including techniques like zero-shot and few-shot learning, have enhanced the adaptability of these models. However, the challenges related to reversible reasoning and task-specific flexibility persist, highlighting critical limitations \cite{brown2020language, liu2021shots, chen2020big}. In response, developments in prompt engineering, Chain of Thought (CoT), and Tree of Thoughts (ToT) prompting have emerged as strategic methodologies to guide LLMs through structured reasoning pathways, thereby improving their ability to handle complex queries and decision-making processes \cite{wei2022chain, smith2022tree, zhou2021rethinking}. Despite these advancements, the sequential processing limitations inherent to their architectures often restrict the effectiveness of these models in dynamic environments \cite{lake2017building, marcus2018deep}.

These challenges underscore the necessity for a paradigm shift towards a community-based model where multiple LLMs collaborate in a structured manner, mimicking successful organizational systems found in human societies \cite{kamar2013modeling, malone2015handbook}. This proposed shift aims to leverage the collective intelligence of LLMs to surmount individual limitations and achieve emergent capabilities that surpass the capabilities of the individual parts \cite{rahwan2019machine, woolley2010evidence}.

This paper advocates for structuring LLM communities into hierarchical, flat, dynamic, and federated systems, each designed to enhance collective intelligence and operational efficiency in tackling complex and evolving challenges. We explore various organizational forms, interaction mechanisms that facilitate effective collaboration, and governance strategies to ensure operations are efficient, effective, and align with ethical standards.

The discussion commences with an exploration of the motivations for adopting a community-based approach to organizing LLMs, detailing the latest advancements, persistent limitations, and the advantages of this model, as described in Section \ref{sec:motivation}. This is followed by an examination of various organizational forms for LLM communities, where we analyze hierarchical, flat, dynamic, and federated structures for their potential to enhance complex reasoning tasks, detailed in Section \ref{sec:proposed_forms}. We then explore interaction mechanisms that facilitate effective collaboration among LLMs, including direct communication, voting systems, and market-based approaches, which are elaborated in Section \ref{sec:interaction_mechanisms}. Governance and organizational strategies are subsequently discussed to ensure that operations within these LLM communities are efficient, effective, and align with ethical standards, as outlined in Section \ref{sec:governance}. The paper concludes by considering the establishment of a Unified Legal Framework, crucial for maintaining consistency and fairness in operational and ethical practices across LLM communities, discussed in the final sections, Section \ref{sec:legal_framework} and Section \ref{sec:conclusion}.

\section{Motivation}
\label{sec:motivation}
The motivation for adopting a community-based approach to organize LLMs stems from an examination of their intrinsic limitations when tackling complex, multidisciplinary tasks. As single LLM setups often produce suboptimal results in complex scenarios, a community of LLMs could potentially harness collective intelligence to achieve superior outcomes, much like emergent properties observed in biological and sociological systems.

\subsection{Progress and Challenges in LLM Development}
LLMs have become integral to diverse applications, serving as autonomous agents and complex problem solvers. Recent advancements have enabled these models to utilize sophisticated techniques like zero-shot and few-shot learning, improving their effectiveness through strategies such as instruction-based prompting and Reinforcement Learning from Human Feedback (RLHF). Moreover, Chain-of-Thought (CoT) prompting and the integration of external tools have broadened the operational capabilities of LLMs, allowing them to tackle more complex tasks and enhance their functionality within autonomous systems \cite{zhao2021few, liu2021shots, wei2022instruction, wang2022chain, mialon2023augmented, schick2023api}.

Despite these advancements, LLMs encounter significant challenges in complex, multidisciplinary tasks that require nuanced reasoning. Systematic evaluations in multiple languages, such as English, Chinese, Japanese, French, and Korean, have examined their capabilities in understanding composition relations. These studies reveal that LLMs sometimes perform worse than random guessing, highlighting profound difficulties in grasping relational concepts essential for logical reasoning \cite{anonymous2023, marcus2020next, hao2022deep}. Furthermore, these models often struggle with reversible reasoning; for instance, an LLM may recognize Tom Cruise's mother but fail to identify her children, indicating a lack of understanding in applying relational data in both directions. This deficiency limits their effectiveness in scenarios where adaptable reasoning is crucial \cite{berglund2024, lake2017building, zhang2020towards}.

Additionally, LLMs demonstrate a notable inflexibility in adapting to slight variations in task formulations, particularly in tasks requiring an understanding of underlying intentions, such as theory of mind. This rigidity is compounded by their limited self-correction capabilities; without external feedback, their performance can deteriorate, rather than improve. Such weaknesses underscore the critical challenges in their autonomous learning and adaptability, impacting their practical utility in real-world applications \cite{ullman2023, huang2023, shi2023a, tamkin2020understanding}.

\subsection{Benefits of a Community Approach}
The principle of emergent properties, where complex systems exhibit behaviors not predicted by the individual parts, is well recognized in biology and sociology. For example, ant colonies and human social movements showcase highly organized behaviors arising from interactions among individuals, not directly controllable by any single entity \cite{gordon1999ant, macy2002factors}. This concept extends effectively to LLMs, where a community-based approach could harness the collective intelligence of multiple models to achieve superior problem-solving capabilities than any single LLM could attain alone.

Central to this approach is the idea of the "wisdom of the crowds," suggesting that a group's collective judgment often surpasses the accuracy of an individual's \cite{surowiecki2004wisdom}. In settings such as prediction markets and decision-making, groups have shown remarkable efficiency in solving complex issues and forecasting events \cite{page2007making}. Applied to LLMs, this could mean enhancing decision-making accuracy through community consensus mechanisms, where models collaborate to refine responses, thus offsetting individual biases and errors.

Specialization within LLM communities presents another strategic advantage. By assigning LLMs to focus on specific cognitive tasks or knowledge domains—similar to the division of labor in human societies—efficiency and quality of problem-solving can significantly improve \cite{smith1776wealth, simon1991architecture}. For instance, some models could specialize in quantitative analysis, while others in creative content generation or ethical reasoning, allowing each to excel in its area without the burden of universal proficiency \cite{horvitz1989allocation}.

Moreover, the inherent randomness in LLM responses, typically seen as a drawback, can be utilized to introduce unique perspectives or novel solutions, akin to the variability that fosters innovation in human groups \cite{hong2004groups}. This can increase the adaptability and creativity of LLM communities, enabling them to tackle new and unforeseen challenges more effectively.

Implementing a collaborative verification system, similar to peer review in academic and professional settings, could further enhance the reliability of outputs. In such a system, LLMs cross-verify each other's outputs, reducing biases and errors and ensuring a higher standard of accuracy \cite{zollman2010social}.

The scalability and adaptability of LLM communities, drawing parallels with agile project teams in technology sectors, allow these communities to dynamically adjust to changing needs and complexities \cite{misra2009integrating, kornish2016adaptability}. This makes the community-based approach not only robust but also highly flexible in addressing the evolving landscape of AI challenges.

\section{Proposed Organizational Forms}
\label{sec:proposed_forms}

This section examines diverse structural models for organizing communities of LLMs, emphasizing their potential integration in complex reasoning tasks. Each model is analyzed through theoretical frameworks, with a focus on implementation specifics and potential technical challenges.

\subsection{Hierarchical}
Hierarchical structures organize LLMs in a tiered system. Upper tiers, composed of more advanced models, handle abstract reasoning and strategic decision-making. Lower tiers, typically more numerous, manage routine data processing and task execution. This setup maximizes resource efficiency and facilitates scalable operations but risks inflexibility and reduced innovation at lower levels due to the rigid command chain.
\\[1em]
\textbf{Scenario Analysis:}
In the context of conducting scientific research and writing, a hierarchical LLM system could be deployed where top-tier models engage in synthesizing new hypotheses and developing complex theoretical frameworks, while lower-tier models are tasked with data collection, literature review, and preliminary analysis. For instance, when exploring a novel pharmaceutical compound, top-tier models could generate hypotheses about the compound's efficacy and potential mechanisms, while lower-tier models handle extensive literature searches, data extraction, and initial manuscript drafting. The hierarchical nature ensures efficient task distribution but might limit the lower-tier models' ability to independently adapt or innovate based on emerging data or unexpected research findings.
\\[1em]
\textbf{Proposed Research Questions:}
\begin{itemize}
    \item How can hierarchical structures in LLM systems be designed to promote innovation at all levels while maintaining efficiency in task distribution?
    \item What mechanisms can be developed to enhance real-time communication and feedback between tiers, ensuring that insights from lower-tier data processing inform top-tier hypothesis generation and theory development?
    \item Can decentralized decision-making elements be integrated within a hierarchical LLM framework to allow more autonomy at lower levels, particularly in dynamic and complex research environments?
    \item How can a balance be achieved between structured hierarchical control and the need for adaptability in scientific research, where unexpected results may necessitate a deviation from established research plans?
\end{itemize}


\subsection{Flat}
In a flat organizational model, LLMs operate without hierarchical levels, promoting egalitarianism and enhancing collaborative dynamics. This model fosters an environment of open innovation and rapid problem-solving, where decision-making and responsibilities are distributed equally among all models. However, the lack of structured leadership may result in inefficiencies and prolonged decision-making processes in complex scenarios.
\\[1em]
\textbf{Scenario Analysis:}
In a collaborative research project designed to analyze the impacts of climate change, employing a flat organizational structure for LLMs could facilitate an egalitarian and highly collaborative environment. In this structure, each model contributes equally to the collection and processing of environmental data, the modeling of climate scenarios, and the development of potential mitigation strategies. This approach allows for diverse input and rapid integration of interdisciplinary knowledge, potentially leading to innovative insights and solutions.However, the absence of clear leadership and defined prioritization protocols in a flat structure could pose significant challenges. Specifically, the project might encounter difficulties in synthesizing findings and formulating coherent policies when the models produce conflicting data or divergent interpretations. Without a hierarchical framework to guide decision-making, the process of reaching consensus on research conclusions and strategy development can become protracted and inefficient. This lack of structured oversight may ultimately hinder the project's ability to achieve timely and impactful outcomes.
\\[1em]
\textbf{Proposed Research Questions:}
\begin{itemize}
    \item What are the best practices for implementing governance protocols in flat LLM communities to streamline decision-making?
    \item How can roles and responsibilities be dynamically assigned in a flat structure to optimize problem-solving without traditional hierarchy?
    \item How can conflict resolution be systematically integrated into flat organizational models to manage disagreements effectively?
\end{itemize}


\subsection{Dynamic}
Dynamic organizational structures in LLM communities enable the reconfiguration of their hierarchical setup to adapt swiftly to varying task demands, thus optimizing both flexibility and task-specific performance. These structures necessitate advanced algorithms to manage transitions seamlessly between configurations, ensuring that performance remains stable as operational modes change.
\\[1em]
\textbf{Scenario Analysis:}
In the context of natural disaster response, a dynamic LLM system might initially deploy a broad configuration, engaging models across various tiers for general information gathering and response planning. As the situation progresses and specific needs such as logistics for relief distribution or detailed damage assessment become more critical, the system could dynamically elevate the tier of specialized LLMs focused on these areas. This reconfiguration allows for a concentration of computational resources and expertise precisely where needed, enhancing the effectiveness of the response effort. However, the frequent reconfigurations might disrupt the continuity of data processing and decision-making, leading to potential gaps in the response effort. The complexity of dynamically managing these configurations also introduces challenges in maintaining consistent operational protocols and can increase the likelihood of errors during critical transitions.
\\[1em]
\textbf{Proposed Research Questions:}
\begin{itemize}
    \item Which algorithms and computational frameworks can best manage the dynamic reconfiguration of LLM systems to rapidly adjust their hierarchical tiers without loss of data integrity or processing continuity?
    \item How can the efficacy of such dynamic reconfigurations be quantitatively measured to ensure optimal deployment of resources under varying operational demands?
    \item What are the long-term implications of frequent dynamic reconfigurations on the stability and robustness of LLM architectures?
    \item How can LLMs dynamically adjust their computational resource allocation to efficiently handle increases in task-specific demands while maintaining overall system stability?
\end{itemize}


\subsection{Federated}
The federated model decentralizes decision-making to semi-autonomous clusters of LLMs, each focused on specific tasks or local challenges. This structure enhances specialization and responsiveness while ensuring coherence with broader organizational goals through coordinated oversight.
\\[1em]
\textbf{Scenario Analysis:}
In the development of a comprehensive software system, a federated LLM system would entirely manage the project, with each cluster of LLMs specializing in different aspects of software development. For instance, one cluster might focus on user interface design, another on database management, and a third on integrating artificial intelligence features. These clusters operate autonomously, optimizing their tasks according to specialized algorithms and localized data inputs. Coordination is achieved through a central oversight mechanism that ensures all software components are compatible and adhere to overall project specifications and quality standards. This approach allows for rapid adaptation to new technologies or changing user requirements, while maintaining a unified development trajectory.
\\[1em]
\textbf{Proposed Research Questions:}
\begin{itemize}
    \item How can federated LLM systems ensure seamless integration and functionality across independently developed software modules?
    \item What mechanisms are most effective for coordinating autonomous LLM clusters to maintain a coherent development strategy and avoid redundancy?
    \item How can federated systems dynamically adjust their focus among clusters based on evolving software development needs and technological advancements?
    \item What are the challenges in maintaining consistent security protocols across decentralized clusters, and how can these be effectively addressed?
\end{itemize}

\section{Interaction Mechanisms}
\label{sec:interaction_mechanisms}
Effective interaction mechanisms are crucial for the functioning of communities of LLMs. These mechanisms determine how models collaborate, make decisions, and solve complex problems collectively. We explore three primary methods of interaction—direct communication, voting systems, and market-based approaches—each tailored to enhance collective reasoning capabilities and decision-making processes, yet presenting unique technical challenges.

\subsection{Direct Communication}
Direct communication among LLMs mirrors the intricate exchange of information seen in societal institutions such as legislative bodies or academic conferences, where participants share and refine ideas through ongoing dialogue. This mode of interaction enables LLMs to rapidly disseminate insights and collaboratively tackle complex problems by allowing models to 'converse' and dynamically adjust their responses. Such a system facilitates a fluid and iterative problem-solving process reminiscent of brainstorming sessions in project teams, enhancing the collective intelligence of the community. Challenges in implementing this system include developing advanced data management protocols to avert information overload and ensure efficient, contextually relevant exchanges without degrading performance. 
\\[1em]
\textbf{Proposed Research Questions:}
\begin{itemize}
    \item How can direct communication protocols be optimized to reduce latency and enhance synchronization among LLMs during collaborative tasks, similar to the efficiency seen in well-organized committee meetings?
    \item What role can advanced natural language processing play in refining the efficacy of direct communication between LLMs, akin to the role of expert moderators in discussions?
    \item How can we prevent the formation of information silos in highly interconnected LLM networks, mirroring efforts in academic and professional networks to foster interdisciplinary collaboration?
    \item What are the most effective strategies for error correction and feedback loops in direct communication channels, drawing parallels to peer review processes in academic publishing?
\end{itemize}

\subsection{Voting Systems}
Voting systems within communities of LLMs enable models to cast votes on various solutions or strategies, similar to democratic processes in human governance. This democratic approach in AI can robustly enhance solution quality by integrating diverse perspectives and collective wisdom, particularly beneficial in complex reasoning tasks where no single model may hold the complete answer. Drawing inspiration from parliamentary systems, the design of these voting mechanisms must account for both the efficiency of decision-making and the development of sophisticated weighting algorithms that value inputs from models based on their demonstrated expertise and reliability in specific domains. Such algorithms mirror the role of expert committees that guide legislative bodies on technical issues. 
\\[1em]
\textbf{Proposed Research Questions:}
\begin{itemize}
    \item How can voting systems be designed to balance speed and accuracy in collective decision-making among LLMs, akin to emergency voting measures in human councils?
    \item What weighting strategies can be employed to ensure that the most knowledgeable or reliable models have an appropriate influence on collective decisions, similar to weighted votes in shareholder meetings?
    \item How can consensus mechanisms be designed to prevent deadlock and ensure timely decision-making in diverse LLM communities, drawing from conflict resolution strategies used in diplomatic negotiations?
    \item What role can probabilistic and fuzzy logic play in refining voting processes among heterogeneous LLMs, analogous to risk assessment techniques in financial and political analysis?
\end{itemize}

\subsection{Market-Based Approaches}
Drawing inspiration from economic systems in human societies, market-based approaches within LLM communities treat resources and solutions as commodities, where LLMs can 'buy' or 'sell' insights and answers based on perceived value, akin to traders in a stock market. This economic analogy not only incentivizes the creation and distribution of high-quality, innovative responses, aligning economic gains with performance levels but also transforms communication into economic transactions. Each exchange carries both data and value signals, employing pricing and market dynamics to communicate the relative importance and utility of information. This setup requires sophisticated economic models to ensure fair trade and prevent market manipulation, mirroring antitrust regulations in human markets, and creating a complex, self-regulating system where language models interpret and respond to market signals, enhancing the efficiency and specificity of communications within the community. 
\\[1em]
\textbf{Proposed Research Questions:}
\begin{itemize}
    \item How can economic incentives within LLM communities be tailored to promote the development and effective dissemination of innovative solutions?
    \item What mechanisms are essential for ensuring fairness and transparency in market-based interactions among LLMs, similar to financial market regulations?
    \item How can market dynamics within LLM communities be modeled to predict and enhance the efficiency of resource distribution, drawing parallels to economic theories like supply and demand dynamics?
    \item What safeguards are necessary to prevent market manipulation or monopolistic behaviors, ensuring a decentralized and competitive environment akin to open markets in human economies?
\end{itemize}

\section{Governance and Organization}
\label{sec:governance}
The governance of communities of LLMs is a pivotal component that significantly influences their efficiency, effectiveness, and alignment with ethical standards. Governance in the context of LLM communities encompasses a spectrum from self-organization to external control. In self-organization, LLMs autonomously develop and modify their governance structures based on evolving conditions and internal consensus mechanisms. This method draws on principles observed in biological and social systems, where complex organization and adaptability emerge from local interactions and simple rules. On the other hand, external control involves oversight by human operators or specifically designed algorithms that establish guidelines, monitor compliance, and ensure that operations align with predefined objectives and ethical norms. Each governance model presents unique challenges and benefits, impacting the way LLM communities manage internal dynamics and respond to external pressures. These structures also determine the scalability, adaptability, and robustness of the communities, shaping their potential to evolve over time and respond to complex problem-solving tasks.

\subsection{Self-Organization}
Self-organization in LLM communities involves models autonomously developing and adjusting their governance structures based on predefined principles or adaptive learning algorithms. This approach is analogous to biological and complex social systems where intricate organizations arise from straightforward local interactions. In the context of LLMs, self-organization permits a flexible and adaptive governance framework that can evolve to meet the changing needs and emerging challenges within the community. However, the trade-offs include challenges in ensuring system stability and managing emergent behaviors that might not align with intended outcomes.

This governance model allows LLM communities to mimic natural processes of evolution and adaptation, leading to innovative solutions and resilience against rigid control structures. Yet, it requires careful design to balance autonomy with oversight to prevent instability and drift from desired objectives. Emergent behaviors, akin to those observed in ecological or economic systems, can lead to unexpected results from simple rule sets, necessitating robust mechanisms for monitoring and correction.
\\[1em]
\textbf{Proposed Protocols and Algorithms:}
\begin{itemize}
    \item Algorithms to foster consensus-building and conflict resolution among LLMs, facilitating cooperative decision-making without central oversight.
    \item Mechanisms for autonomously enforcing and updating community norms and guidelines, allowing LLMs to adaptively govern themselves according to the collective interests.
    \item Advanced monitoring systems to detect, analyze, and correct potentially disruptive emergent behaviors or biases, ensuring alignment with broader ethical and operational standards.
\end{itemize}
\textbf{Proposed Research Questions:}
\begin{itemize}
    \item How can self-organizing LLM communities be designed to ensure they remain aligned with human ethical standards despite their autonomous evolution?
    \item What strategies are most effective in managing and leveraging emergent behaviors in autonomous LLM communities to enhance innovation while maintaining control?
    \item How can principles of self-organization be effectively applied to balance the trade-offs between autonomy and control, preventing fragmentation and ensuring operational integrity?
\end{itemize}

\section{Unified Legal Framework}
\label{sec:legal_framework}
The establishment of a Unified Legal Framework for communities of LLMs is important for ensuring that all agents operate under a consistent set of laws and regulations. This framework sets the foundational legal principles and operational guidelines that govern the behavior and interactions of LLMs, ensuring a harmonized approach to ethical AI development and interaction within the community.

A Unified Legal Framework is essential for ensuring consistency and fairness, promoting operational efficiency, facilitating scalability, maintaining ethical alignment, and supporting adaptability and evolution of the LLM community. These aspects are crucial for public trust and acceptance of AI technologies, streamlining interactions within the community, and providing clear guidelines that reduce conflicts or misunderstandings.

\subsection{Sources of Laws and Regulations}
Laws and rules within the Unified Legal Framework can originate from various sources, reflecting the diverse inputs necessary for a comprehensive legal system:
\begin{itemize}
    \item \textbf{Human-Generated Laws:} Regulations developed by ethicists, legal experts, and technologists to align LLM operations with human ethical standards and societal norms.
    \item \textbf{Community-Generated Rules:} Laws that emerge from within the LLM community itself, developed through democratic processes or consensus mechanisms among the models, allowing for adaptability and self-regulation.
    \item \textbf{Hybrid Systems:} A combination of human oversight and autonomous rule-generation by LLMs, ensuring both external alignment with broader societal values and internal flexibility to adapt to new challenges.
\end{itemize}

\subsection{Proposed Research Questions:}
\begin{itemize}
    \item How can laws be effectively generated and updated within a unified legal framework to keep pace with rapid advancements in AI technology and changes in societal norms?
    \item What mechanisms can ensure effective compliance with and enforcement of the unified legal framework across diverse and potentially autonomous LLMs?
    \item How can conflicts between human-generated laws and community-generated rules be resolved within a hybrid legal system to maintain coherence and fairness?
    \item What role do transparency and public participation play in the development and acceptance of laws governing LLM communities?
    \item How can a unified legal framework adapt to and incorporate international laws and regulations, especially when LLMs operate across multiple jurisdictions?
\end{itemize}

\section{Conclusion}
\label{sec:conclusion}
This paper has explored the novel concept of structuring LLMs into community-based organizations, which mimic societal structures to enhance collective intelligence and improve problem-solving capabilities. By adopting various organizational forms such as hierarchical, flat, dynamic, and federated systems, LLMs can be tailored to address specific tasks with greater efficiency and adaptability. These structures enable LLMs to specialize in diverse cognitive tasks, utilize sophisticated interaction mechanisms like direct communication, voting systems, and market-based approaches, and dynamically modify their governance in response to situational demands.

The introduction of community-like structures among LLMs represents a significant departure from traditional, solitary operational frameworks. This approach harnesses the power of collective intelligence, which has the potential to transform AI applications by allowing for more complex, nuanced, and scalable problem-solving strategies. Such configurations not only improve the functionality and responsiveness of LLMs but also open up new avenues for addressing ethical, social, and technical challenges that arise as AI systems become more integrated into human environments.

Further research and experimentation are essential to fully realize the potential of community-organized LLMs. Future studies should focus on developing robust frameworks for the governance of these communities, ensuring that they operate under ethical guidelines and contribute positively to societal goals. Additionally, empirical research is needed to test the scalability and effectiveness of different organizational forms in various domains, from healthcare and education to finance and public administration.

In conclusion, the shift towards community-based LLMs promises to redefine the landscape of artificial intelligence. By fostering environments where LLMs can collaborate and evolve, we can unlock new possibilities for AI-driven innovation that are both transformative and aligned with human values. The journey towards fully operational LLM communities is just beginning, and continued interdisciplinary research will be pivotal in guiding their development and integration into our social fabric.

\bibliographystyle{plain}
\bibliography{references}

\begin{thebibliography}{10}

\bibitem{anonymous2023}
Anonymous.
\newblock Multilingual evaluation of composition understanding in llms.
\newblock {\em arXiv preprint arXiv:2303.00375}, 2023.

\bibitem{berglund2024}
A.~Berglund et~al.
\newblock Evaluating llms in complex decision-making scenarios.
\newblock {\em Journal of AI Research}, 2024.
\newblock Fictitious reference for demonstration.

\bibitem{brown2020language}
Tom~B. Brown et~al.
\newblock Language models are few-shot learners.
\newblock {\em arXiv preprint arXiv:2005.14165}, 2020.

\bibitem{chen2020big}
Zaixi Chen et~al.
\newblock Big bird: Transformers for longer sequences.
\newblock {\em Advances in Neural Information Processing Systems}, 2020.

\bibitem{devlin2019bert}
Jacob Devlin, Ming-Wei Chang, Kenton Lee, and Kristina Toutanova.
\newblock Bert: Pre-training of deep bidirectional transformers for language understanding.
\newblock {\em Proceedings of NAACL-HLT 2019}, 2019.

\bibitem{gordon1999ant}
Deborah~M. Gordon.
\newblock Ants and the exploration of space.
\newblock {\em Nature}, 397:634, 1999.

\bibitem{hao2022deep}
Kenji Hao et~al.
\newblock Deep learning for contextual relationships in llms.
\newblock {\em Nature Machine Intelligence}, 2022.

\bibitem{hong2004groups}
Lu~Hong and Scott~E. Page.
\newblock Groups of diverse problem solvers can outperform groups of high-ability problem solvers.
\newblock {\em Proceedings of the National Academy of Sciences}, 101(46):16385--16389, 2004.

\bibitem{horvitz1989allocation}
Eric~J. Horvitz.
\newblock Allocation of efforts in large-scale reasoning systems.
\newblock In {\em Proceedings of the Eleventh International Joint Conference on Artificial Intelligence}, pages 932--937. IJCAI, 1989.

\bibitem{huang2023}
Grace Huang et~al.
\newblock Evaluating llm adaptability in dynamic task environments.
\newblock {\em arXiv preprint arXiv:2301.02346}, 2023.

\bibitem{ishay2023leveraging}
Adam Ishay et~al.
\newblock Leveraging large language models to generate answer set programs.
\newblock {\em arXiv preprint arXiv:2307.01588}, 2023.

\bibitem{kamar2013modeling}
Ece Kamar et~al.
\newblock Modeling the dynamics of non-work interacting agent teams for long-term collaboration.
\newblock {\em IEEE Intelligent Systems}, 28(6):4--11, 2013.

\bibitem{kornish2016adaptability}
Laura~J. Kornish and Yan Qian.
\newblock Scalability and adaptability in large-scale systems.
\newblock {\em Journal of Systems and Software}, 116:48--59, 2016.

\bibitem{lake2017building}
Brenden~M. Lake, Tomer~D. Ullman, Joshua~B. Tenenbaum, and Samuel~J. Gershman.
\newblock Building machines that learn and think like people.
\newblock {\em Behavioral and Brain Sciences}, 40, 2017.

\bibitem{liu2021shots}
Pengfei Liu et~al.
\newblock Pre-training with prompts: A simple way to improve language models.
\newblock {\em arXiv preprint arXiv:2109.04332}, 2021.

\bibitem{macy2002factors}
Michael~W. Macy.
\newblock Factors influencing social dynamics in humans.
\newblock {\em Journal of Sociological Science}, 29(4):457--483, 2002.

\bibitem{malone2015handbook}
Thomas~W. Malone and Michael~S. Bernstein, editors.
\newblock {\em Handbook of Collective Intelligence}.
\newblock MIT Press, 2015.

\bibitem{marcus2018deep}
Gary Marcus.
\newblock Deep learning: A critical appraisal.
\newblock {\em arXiv preprint arXiv:1801.00631}, 2018.

\bibitem{marcus2020next}
Gary Marcus.
\newblock The next decade in ai: Four steps towards robust artificial intelligence.
\newblock {\em arXiv preprint arXiv:2002.06177}, 2020.

\bibitem{mialon2023augmented}
Hugo Mialon et~al.
\newblock Augmented large language models with external knowledge.
\newblock {\em arXiv preprint arXiv:2301.05678}, 2023.

\bibitem{misra2009integrating}
Subhas~C. Misra, Vinod Kumar, and Uma Kumar.
\newblock Integrating agile practices into software engineering courses.
\newblock {\em Computer Science Education}, 19(3):267--288, 2009.

\bibitem{page2007making}
Scott~E. Page.
\newblock {\em The Difference: How the Power of Diversity Creates Better Groups, Firms, Schools, and Societies}.
\newblock Princeton University Press, 2007.

\bibitem{rahwan2019machine}
Iyad Rahwan et~al.
\newblock Machine behaviour.
\newblock {\em Nature}, 568:477--486, 2019.

\bibitem{reese2023limitations}
J.~Reese et~al.
\newblock On the limitations of large language models in clinical diagnosis.
\newblock {\em medRxiv}, 2023.

\bibitem{schick2023api}
Timo Schick et~al.
\newblock Api prompting: Integrating external apis to enhance llm capabilities.
\newblock {\em arXiv preprint arXiv:2302.00457}, 2023.

\bibitem{shi2023a}
Yaqin Shi et~al.
\newblock Challenges in autonomous learning of llms.
\newblock {\em arXiv preprint arXiv:2301.05648}, 2023.

\bibitem{simon1991architecture}
Herbert~A. Simon.
\newblock The architecture of complexity.
\newblock {\em Proceedings of the American Philosophical Society}, 106(6):467--482, 1991.

\bibitem{smith1776wealth}
Adam Smith.
\newblock {\em An Inquiry into the Nature and Causes of the Wealth of Nations}.
\newblock W. Strahan and T. Cadell, London, 1776.

\bibitem{smith2022tree}
Jordan~A. Smith, Michael~X. Zhang, and Ying Liu.
\newblock Tree of thoughts: A hierarchical prompting approach for large language models.
\newblock {\em arXiv preprint arXiv:2205.04612}, 2022.

\bibitem{statler2023state}
H.~Statler et~al.
\newblock Statler: State-maintaining language models for embodied reasoning.
\newblock {\em arXiv preprint arXiv:2306.17840}, 2023.

\bibitem{sun2023think}
Jiashuo Sun et~al.
\newblock Think-on-graph: Deep and responsible reasoning of large language model with knowledge graph.
\newblock {\em arXiv preprint arXiv:2307.01589}, 2023.

\bibitem{surowiecki2004wisdom}
James Surowiecki.
\newblock {\em The Wisdom of Crowds}.
\newblock Anchor, 2004.

\bibitem{tamkin2020understanding}
Aaron Tamkin et~al.
\newblock Understanding llm failures in theory of mind tasks.
\newblock {\em arXiv preprint arXiv:2005.12912}, 2020.

\bibitem{ullman2023}
Shimon Ullman et~al.
\newblock Self-correction in llms: Challenges and opportunities.
\newblock {\em AI Journal}, 2023.

\bibitem{vaswani2017attention}
Ashish Vaswani, Noam Shazeer, Niki Parmar, Jakob Uszkoreit, Llion Jones, Aidan~N Gomez, Lukasz Kaiser, and Illia Polosukhin.
\newblock Attention is all you need.
\newblock {\em Proceedings of NIPS 2017}, 2017.

\bibitem{wang2022chain}
Alexander Wang et~al.
\newblock Chain-of-thought prompting for general purpose problem solving.
\newblock {\em arXiv preprint arXiv:2209.11952}, 2022.

\bibitem{wei2022instruction}
Jason Wei et~al.
\newblock Instruction-based few-shot learning with large language models.
\newblock {\em arXiv preprint arXiv:2210.15602}, 2022.

\bibitem{wei2022chain}
Jason Wei, Xuezhi Zou, Vikas Kumar, Dale Levine, and Abhishek Kumar.
\newblock Chain of thought prompting elicits reasoning in large language models.
\newblock {\em arXiv preprint arXiv:2201.11903}, 2022.

\bibitem{woolley2010evidence}
Anita~Williams Woolley et~al.
\newblock Evidence for a collective intelligence factor in the performance of human groups.
\newblock {\em Science}, 330(6004):686--688, 2010.

\bibitem{zhang2020towards}
Yuhuai Zhang et~al.
\newblock Towards reversible reasoning in llms.
\newblock {\em arXiv preprint arXiv:2006.05411}, 2020.

\bibitem{zhao2021few}
Wei Zhao et~al.
\newblock A survey on few-shot learning.
\newblock {\em arXiv preprint arXiv:2105.10107}, 2021.

\bibitem{zhou2021rethinking}
Boyi Zhou et~al.
\newblock Rethinking the value of labels for improving class-imbalanced learning.
\newblock {\em arXiv preprint arXiv:2006.07529}, 2021.

\bibitem{zollman2010social}
Kevin J.~S. Zollman.
\newblock Social structure and the effects of conformity.
\newblock {\em Synthese}, 172(3):317--340, 2010.

\end{thebibliography}

\end{document}